\documentclass[letterpaper, 10 pt, conference]{ieeeconf}  

\IEEEoverridecommandlockouts                              

\title{\LARGE \bf
OASIS: Optimal Arrangements for Sensing in SLAM
}

\author{Pushyami Kaveti$^{1}$, Matthew Giamou$^{2}$, Hanumant Singh$^{1}$, and David M. Rosen$^{1}$
\thanks{$^{1}$P. Kaveti, H. Singh, and D. Rosen are with Northeastern University, USA {\tt\scriptsize \{kaveti.p, ha. singh, d.rosen\} @northeastern.edu}}. \thanks{$^{2}$ M. Gaimou is with McMaster University, CA {\tt\scriptsize giamoum@mcmaster.ca}}%
}
\usepackage{amsmath} 
\usepackage{amssymb}
\usepackage{amsthm} 

\usepackage{graphics} 
\usepackage{epsfig} 
\DeclareMathAlphabet{\mathcal}{OMS}{cmsy}{m}{n} 

\usepackage{multicol}
\usepackage[bookmarks=true]{hyperref}

\usepackage{textcomp}

\usepackage{xparse} 
\usepackage{amsfonts} 
\usepackage{dsfont}
\usepackage{graphicx}
\usepackage{url}
\usepackage{booktabs}
\usepackage[para]{threeparttable}
\usepackage{multirow}
\hypersetup{colorlinks,linkcolor={blue},citecolor={green},urlcolor={blue}}  

\usepackage{xcolor}
\usepackage{framed}
\usepackage{caption}
\usepackage{varwidth}
\usepackage{subcaption}
\usepackage[utf8]{inputenc}
\usepackage{pgfplots}
\usepackage{adjustbox}
\DeclareUnicodeCharacter{2212}{−}
\usepgfplotslibrary{groupplots,dateplot}

\usepackage{tikz}

\usetikzlibrary{patterns,shapes.arrows}
\usetikzlibrary{shapes.geometric}
\pgfplotsset{compat=newest}
\usepackage{balance}

\usepackage{pifont}
\usepackage{algorithm}
\usepackage{algpseudocode}
\makeatletter
\let\OldStatex\Statex
\renewcommand{\Statex}[1][0]{%
  \setlength\@tempdima{\algorithmicindent}%
  \OldStatex\hskip\dimexpr#1\@tempdima\relax}
\makeatother

\algnewcommand\algorithmicinput{\textbf{Input:}}
\algnewcommand\Input{\item[\algorithmicinput]}
\algnewcommand\algorithmicoutput{\textbf{Output:}}
\algnewcommand\Output{\item[\algorithmicoutput]}

\usepackage{cleveref}

\usepackage{xspace}
\usepackage{setspace}
\colorlet{shadecolor}{gray!10}
   
\newtheorem{problem}{Problem}

\newtheorem{proposition}{Proposition}

\NewDocumentCommand\bbm{}{ \begin{bmatrix} }
\NewDocumentCommand\ebm{}{ \end{bmatrix} }

\NewDocumentCommand\Schur{}{ \mathrm{Schur} }

\NewDocumentCommand\MinEig{}{ \lambda_1 }

\NewDocumentCommand\ArgMin{m}{ \operatorname*{argmin}_{#1} }

\NewDocumentCommand\Cardinality{m}{ \left\vert#1\right\vert }

\NewDocumentCommand\Sym{}{ \mathbb{S} }
\NewDocumentCommand\SymmetricMatrices{m}{\Sym^{#1}}
\NewDocumentCommand\PSDMatrices{m}{\Sym^{#1}_+}

\NewDocumentCommand\BaseSet{}{ \mathcal{S} }

\NewDocumentCommand\SetFunction{}{ f }

\NewDocumentCommand\LieGroupSE{m}{ \mathrm{SE}(#1) }

\NewDocumentCommand\Cov{m}{\operatorname{Cov}{#1}}

\NewDocumentCommand\Information{}{ \mathcal{I} }

\NewDocumentCommand\AbsoluteValue{m}{ \Cardinality{m} }

\NewDocumentCommand\Estimate{m}{\hat{#1}}

\NewDocumentCommand\Optimal{m}{{#1}^*}

\NewDocumentCommand\Inv{m}{{#1}^{-1}}

\NewDocumentCommand\Defined{}{\triangleq}

\NewDocumentCommand\Landmark{}{ l }
\NewDocumentCommand\LandmarkSet{}{ L }
\NewDocumentCommand\Pose{}{ x }
\NewDocumentCommand\PoseSet{}{ X }

\NewDocumentCommand\NumCameras{}{ K }
\NewDocumentCommand\NumCandidates{}{ N }
\NewDocumentCommand\NumLandmarks{}{ N_l }
\NewDocumentCommand\NumPoses{}{ N_x }
\NewDocumentCommand\NumMeasurements{}{ M }
\NewDocumentCommand\Measurement{}{\tilde{z}}
\NewDocumentCommand\MeasurementSet{}{\tilde{Z}}
\NewDocumentCommand\Objective{}{ f_{\mathrm{E}} }
\begin{document}

\maketitle
\thispagestyle{empty}
\pagestyle{empty}

\begin{abstract}

The number and arrangement of sensors on mobile robot dramatically influence its perception capabilities. Ensuring that sensors are mounted in a manner that enables accurate detection, localization, and mapping is essential for the success of downstream control tasks. However, when designing a new robotic platform, researchers and practitioners alike usually mimic standard configurations or maximize simple heuristics like field-of-view (FOV) coverage to decide where to place exteroceptive sensors. In this work, we conduct an information-theoretic investigation of this overlooked element of robotic perception in the context of simultaneous localization and mapping (SLAM). We show how to formalize the sensor arrangement problem as a form of subset selection under the E-optimality performance criterion. While this formulation is NP-hard in general, we show that a combination of greedy sensor selection and fast convex relaxation-based post-hoc verification enables the efficient recovery of \emph{certifiably optimal} sensor designs in practice. Results from synthetic experiments reveal that sensors placed with OASIS outperform benchmarks in terms of mean squared error of visual SLAM estimates.

\end{abstract}

\section{Introduction} \label{sec:introduction}

Advances in sensing apparatus often drive improvements in robotic perception.  For example, the availability of lidar sensors that produce accurate and dense point clouds has enabled substantial progress with self-driving vehicles.
Similarly, the proliferation of lightweight, low-power, and affordable sensors for mobile phones and other consumer electronics has provided researchers with a growing arsenal of sensing modalities with which to outfit mobile robots.
However, in spite (or perhaps because) of this perceptual embarrassment of riches, there is at present no definitive theoretical framework for understanding how sensors should be mounted on a mobile robot designed for some task. 

In this paper, we address this significant research gap by developing a formal methodology to optimally arrange the sensors on a robot designed to perform SLAM.
While sensor placement also concerns other tasks like tracking and collision detection, we focus on SLAM as it is a fundamental capability that significantly impact a robot's capacity to perform other downstream tasks. 
%

Our proposed method, \emph{Optimal Arrangements for Sensing in SLAM} (OASIS), can be applied to optimize the design of any mapping or navigation system that fuses independent measurements from multiple sensors, and whose operation in representative environments can be easily simulated.  In brief, our method accepts as input a finite (but potentially large) list of candidate sensor mountings, and selects a subset that maximizes an information-theoretic measure of localization accuracy evaluated on a sample set of simulated robot trajectories.
%
%
If a robot is expected to take a wide range of trajectories across many environments, a large number of heterogeneous scenarios can be simulated to find a sensor arrangement that performs well on average.  Alternatively, our method can also benefit a robot constrained to specific paths in a fixed environment (e.g.\ as in warehouse operations) to find an arrangement that is tailored to its specific niche.
%
\begin{figure}
\centering
\captionsetup{font={footnotesize}, labelfont={bf,sf}}
\includegraphics[scale=0.35]{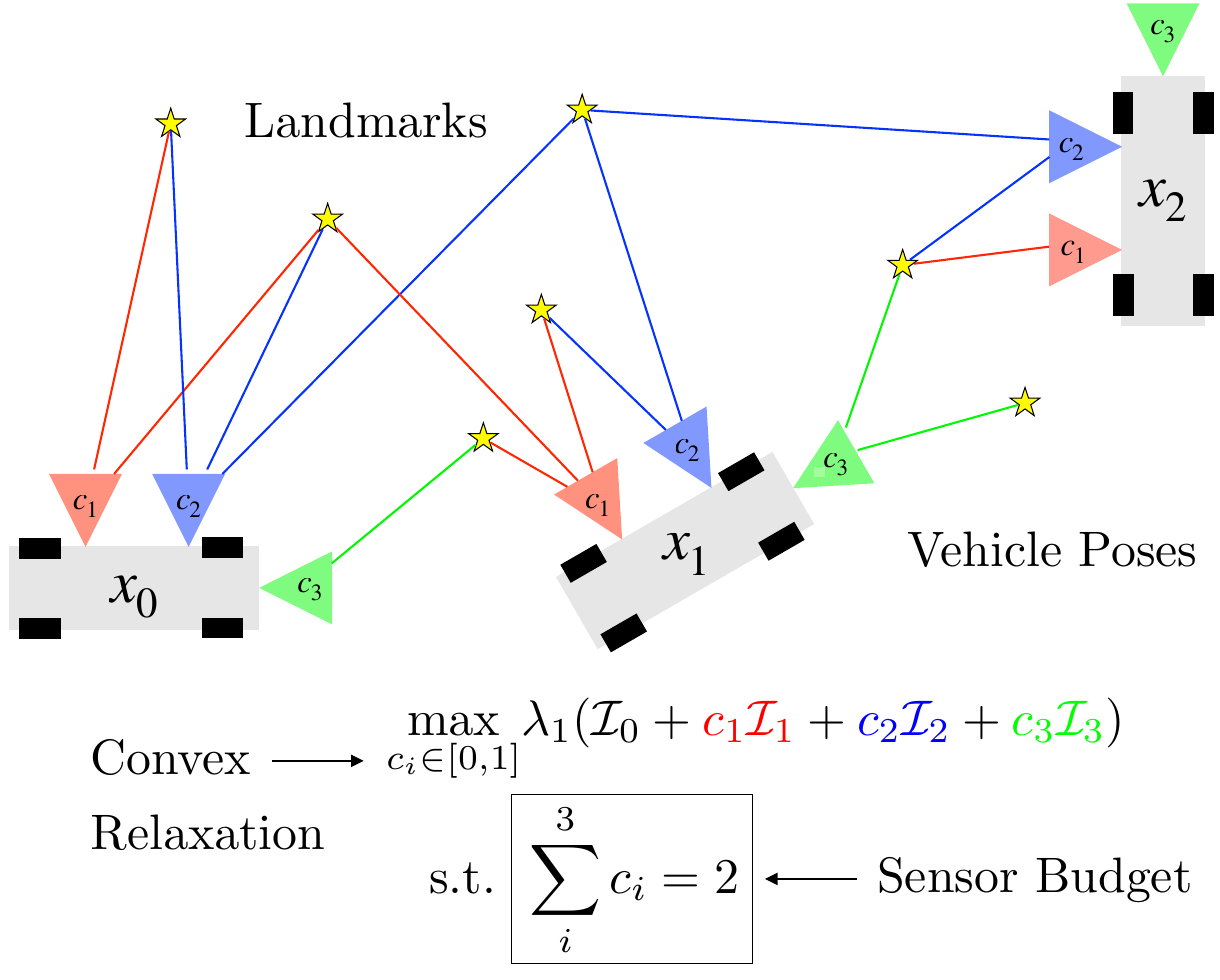}
\caption{Overview of OASIS. 
		 At each pose $\Pose_i$, sensor $c_j$ observes some subset of the landmarks.
		 OASIS maximizes the minimum eigenvalue of the joint Fisher information matrix, which is composed of sub-matrices $\mathcal{I}_j$ from each sensor $c_j$.
		 In this example, the sensor ``budget" limits us to selecting two out of the three candidate sensors.
		 Finally, note that the discrete binary variables indicating sensor selection have been \emph{relaxed} to a convex superset.} \label{fig:overview}
\vspace{-3mm}
\end{figure}
OASIS has three key components that make it unique:
\begin{enumerate}
    \item a design space consisting of a finite set of candidate sensor mountings, leading to a set function maximization problem that can be expressed as a binary integer program (IP); 
	\item a computationally tractable objective function based on E-optimality (i.e.\ minimum eigenvalue maximization) of a Schur complement of the Fisher information matrix of landmark-based SLAM; and
	\item an efficient optimization approach that combines greedy sensor selection with a convex relaxation-based computation of an upper bound to verify optimality. 
\end{enumerate} 
As we will see in \Cref{sec:related_work}, OASIS provides roboticists with a general and easy-to-use tool for designing sensor rigs tailored to accurate SLAM.
Our experiments in \Cref{sec:experiments} demonstrate that OASIS scales gracefully to large numbers of candidate sensor mountings, while recovering sensor designs that are essentially optimal. The code and datasets are open sourced\footnote{ \label{note1} \scriptsize \url{https://github.com/PushyamiKaveti/optimal\_camera\_placement}}.

\section{Related Work} \label{sec:related_work}
In this section we summarize prior work on the design of sensor architectures for mobile robots, and set function optimization methods for perception tasks similar to ours.

\subsection{Design of Perception Systems}
Optimal sensor placement is a well-explored topic in the context of placing sensors in fixed positions in an environment \cite{morsly2009optimal}\cite{nikolaidis2009optimal}\cite{mavrinac2013modeling}. However, sensor design on mobile agents has received relatively limited attention. Recent studies in this area differ from our approach in either the objective function being optimized or the specific downstream target task. Most methods aim to place sensors to maximize \emph{coverage} of an environment ~\cite{zhao_approximate_2013, nikolaidis2009optimal, indu2020optimal, brandao2020placing, liu2019should}, but there is relatively little work on the optimal design of \emph{multi-sensor perception systems} for mobile robot localization and mapping.

Recent research has examined the impact of stereo camera orientation on the accuracy of visual odometry(VO) for aerial \cite{kelly2007experimental} and ground vehicles \cite{peretroukhin2014optimizing}, favoring oblique camera angles to the direction of travel for performance improvement.  Other work \cite{manderson2018gaze} uses a pan-tilt unit to adjust the "gaze" of a stereo camera toward salient features, enabling adaptability to various motion profiles and environments. Lastly, an observability analysis of stereo radar placement is experimentally validated in \cite{corominas-murtra_observability_2016}. In contrast to these works, we address the placement of multiple generic sensors, utilizing an information-theoretic optimality criterion for SLAM.


In SLAM, information-theoretic objectives are commonly applied in active SLAM \cite{sim2005global}, feature selection \cite{zhao_good_2018}, and dynamic sensor selection \cite{chiu2014constrained}. Previous works that have explored information-theoretic sensor placement deal with bespoke sensors like radar~\cite{corominas-murtra_observability_2016}, localization with fixed sensors ~\cite{szaloki_camera_2015}, or object detection tasks~\cite{roos2022method}. The recent literature on \emph{co-design} of complex systems presents a framework for optimization with highly-coupled multidisciplinary constraints and objectives~\cite{censi2016monotone}. This approach has been applied to the Pareto-efficient design of autonomous drones~\cite{zardini2021co} and self-driving vehicles~\cite{zardini_codesign_2021}, including sensor placement on an autonomous vehicle considering candidates formed by discretizing yaw angle~\cite{collin2020multiobjective}. Our method diverges from multi-disciplinary design criteria and instead focuses on a much finer discretization of $\LieGroupSE{3}$ to generate a comprehensive set of candidate sensor mountings. Furthermore, our approach replaces the common (but computationally expensive) D-optimality performance criterion with the (much more tractable) E-optimality criterion.

\subsection{Set Function Optimization for Perception}   
Set function optimization is a natural formulation for many problems, but finding optimal solutions to these challenging combinatorial problems is NP-hard in general. The theory of submodular function maximization provides tractable solutions with \textit{a priori} suboptimality guarantees~\cite{krause_submodular_2014}.
Alternatively, convex relaxations can also provide fast approximate solutions and upper bounds on the optimal value. 

Set function maximization with submodularity-based guarantees is used in \cite{zhang_good_2015} with an observability metric to greedily select visual features that contribute to accurate solutions of SLAM.
In \cite{zhao_good_2018}, a similar approach is employed with information-theoretic criteria for monocular visual SLAM.
This efficient feature selection strategy is extended in \cite{carlone2018attention} to visual-inertial navigation which simulates a short horizon of robot dynamics to \emph{anticipate} future motion and its effect on  feature importance.

Set function maximization has also been employed in the pose graph optimization (PGO) formulation of SLAM, especially in multi-agent settings.
In \cite{khosoussi_reliable_2019}, graph-theoretic properties of PGO are shown to be intimately related to estimation quality and can be exploited to design greedy measurement selection algorithms with suboptimality guarantees or convex relaxations.
A more efficient approach to graph pruning for PGO is used in \cite{doherty2022spectral}: instead of the costly D-criterion of \cite{khosoussi_reliable_2019}, the E-criterion is approximately minimized with the Frank-Wolfe method. 
A related sparsification strategy is applied to the problem of resource-constrained cooperative SLAM (CSLAM) in \cite{giamou2018talk}.
Depending on the choice of optimality criterion, suboptimality guarantees exist for greedy solutions to this ``resource-aware" formulation of CSLAM~\cite{tian2018near, tian_resourceaware_2021}.

In \cite{kayhani2023perception}, the sensors mounted on autonomous agents are considered constant parameters while the poses of fiducial markers in an indoor construction environment are instead treated as design variables. 
Much like the sensor selection literature, a discrete set of candidate marker poses is considered, and an information-theoretic cost function is used.
However, a genetic algorithm without any optimality guarantees is used to generate an optimal tag selection.

Our approach is most closely related to the sensor architecture design problem described in \cite{collin_resilient_2019}, which considers multi-objective optimization for a heterogeneous set of sensors and relatively few candidate mountings. 
OASIS is novel in that it selects a sensor arrangement that is frequently certifiably information-theoretically optimal, even when applied to a large number of candidate sensor mountings.

\section{Problem Statement} \label{sec:problem_statement}


In this section, we show how to formulate the optimal sensor arrangement problem for SLAM as an integer program (IP) with binary variables. 
%
%

\subsection{Parameterizing the space of sensor arrangements}
In order to ``search" for optimal sensor arrangements, we must first parameterize the space of possible designs.  Given a model of a robot chassis, a natural approach might be to first decide upon the number and type of sensors to select, and then determine {where} on the robot to mount them.  While intuitively appealing, this approach leads to a very challenging nonconvex problem that requires joint optimization over a product of (discrete) \emph{sensor selection} variables and (continuous) \emph{sensor pose} variables. 

Instead, we propose to formulate the \emph{sensor arrangement} problem as an instance of \emph{subset selection}.  Specifically, we assume that we are given a finite set $\BaseSet$ whose elements completely enumerate all possible \emph{sensor mountings} (i.e.\ a decision to mount a \emph{specific sensor} at a \emph{specific pose} on the robot chassis).  Designing a sensor arrangement then amounts to selecting a \emph{subset} $S \subset \BaseSet$ of specific sensor mountings.

Our formulation provides several advantages.  First, it straightforwardly captures the fact that, due to mechanical constraints, most robot chassis have only a finite set of locations on which sensors may be mounted.  Second, it avoids the need to perform joint optimization over discrete (sensor selection) and continuous (sensor pose) variables; instead, our approach models the set of possible sensor poses by \emph{enumerating} a discrete set of distinct sensor mounting candidates in $\BaseSet$.  While this leads to an increase in the number of decision variables versus a hybrid discrete-continuous formulation (whose magnitude depends upon how finely the set of candidate sensor poses is discretized), our approach has the advantage that (as we show in the sequel) we can leverage fast approximate combinatorial optimization algorithms to {very} efficiently recover high-quality solutions from the set of candidate mountings $\BaseSet$, even when this set is large.

\subsection{Modeling SLAM performance of sensor arrangements} \label{sec:SLAM}

In this section we describe how to model the SLAM performance of a given sensor arrangement.

Let us imagine our mobile robot navigating through an initially unknown environment containing $\NumLandmarks$ uniquely recognizable landmarks $\LandmarkSet \triangleq \lbrace \Landmark_i \rbrace_{i=1}^{\NumLandmarks}$.  As the robot explores, it moves through a sequence of poses $\PoseSet = \lbrace \Pose_i \rbrace_{i=1}^{\NumPoses} \subset \LieGroupSE{d}$ while collecting measurements from its onboard sensors.  Let us denote by $\MeasurementSet = \lbrace \Measurement_i \rbrace_{i=1}^\NumMeasurements$ the complete set of measurements generated by \emph{all} candidate sensor mountings in $\BaseSet$.  We assume that each of these is sampled \emph{independently} from a known sensor model of the form:\footnote{In general each measurement $\Measurement_i$ will only \emph{explicitly} depend upon a small \emph{subset} of the latent states $(\PoseSet, \LandmarkSet)$.  While exploiting this fact is essential for solving SLAM problems efficiently in practice (as in e.g.\ factor-graph-based SLAM \cite{Dellaert2017Factor}), these details are unimportant for understanding the derivation of OASIS, and so we elide them here.}
\begin{equation}
\label{measurement_model}
\Measurement_i \sim p_i(\cdot \mid \PoseSet, \LandmarkSet) \quad i \in [\NumMeasurements].
\end{equation}

Now let us consider our robot's SLAM performance under a candidate sensor arrangement $S \subset \BaseSet$.  Without loss of generality, letting $\NumCandidates \triangleq \lvert \BaseSet \rvert$, we may label the candidates in $\BaseSet$ by $1, \dotsc, \NumCandidates$, and then \emph{identify} each subset $S \subset \BaseSet$ with a binary vector $s \in \lbrace 0, 1 \rbrace^{\NumCandidates}$ defined by:
\begin{equation}
s_i = 
\begin{cases}
1, & i \in S, \\
0, & i \notin S.
\end{cases}
\end{equation}
Similarly, let us denote by $\ell \colon [\NumMeasurements] \to \NumCandidates$ the function that assigns to each $i \in [\NumMeasurements]$ the label of the sensor mounting that produced the $i$th measurement $\Measurement_i$.  Using this notation, we can parameterize the joint likelihood of the data that would be available to our robot under sensor arrangement $s$ as:
\begin{equation}
\label{parameterized_SLAM_joint_likelihood}
p(\MeasurementSet | \PoseSet, \LandmarkSet; s) \triangleq \prod_{i= 1}^{\NumMeasurements} p_i(\Measurement_i \mid \PoseSet, \LandmarkSet)^{s_{\ell(i)}}.
\end{equation}
In turn, the specific instance of the SLAM maximum-likelihood estimation that our robot would solve under the measurements afforded by arrangement $s$ is:
\begin{equation}
\label{parameterized_SLAM_MLE}
(\Estimate{\PoseSet}(s), \Estimate{\LandmarkSet}(s)) = \ArgMin{\PoseSet, \LandmarkSet} \left \lbrace \sum_{i=1}^\NumMeasurements - s_{\ell(i)} \log p_i(\Measurement_i \mid \PoseSet, \LandmarkSet) \right \rbrace.
\end{equation}

\subsection{Fisher Information and the Cram{\'e}r-Rao Lower Bound}
Recall that the \emph{Cram{\'e}r-Rao Lower Bound} (CRLB) provides a \emph{lower bound} (in the Loewner order sense) on the achievable covariance of any unbiased maximum-likelihood estimator $\Estimate{\Theta}$ of a fixed but unknown parameter $\Theta$ \cite{Ferguson1996Course}:
\begin{equation}
\label{CRLB}
\Cov{\Estimate{\Theta}} \succeq  \Inv{\Information(\Theta)},
\end{equation}
where the matrix $\Information(\Theta)$ appearing on the right-hand side of \eqref{CRLB} is the \emph{Fisher information matrix} (FIM):
\begin{equation}
\Information(\Theta) \triangleq \mathbb{E}_{\MeasurementSet}\left[- \frac{\partial^2}{\partial \Theta^2} \log p(\MeasurementSet \mid \Theta) \right].
\end{equation}

For the SLAM likelihood \eqref{parameterized_SLAM_joint_likelihood}, the CRLB takes the form:
\begin{equation}
\label{parameterized_SLAM_information_matrix}
\Information(\PoseSet, \LandmarkSet; s) = \sum_{i = 1}^{\NumMeasurements} s_{\ell(i)}\mathbb{E}_{\Measurement_i} \left[- \frac{\partial^2}{\partial^2 (\PoseSet, \LandmarkSet)} \log p(\Measurement_i \mid \PoseSet, \LandmarkSet) \right].
\end{equation}
Note that the conditional independence of the measurements $\Measurement_i$ in \eqref{measurement_model} implies that $\Information(\PoseSet, \LandmarkSet; s)$ is the \emph{sum} of the information matrices contributed by \emph{each individual observation} $\Measurement_i$ that is available under design $s$.  Equivalently, writing:
\begin{equation}
\Information_k(\PoseSet, \LandmarkSet) \triangleq \sum_{\lbrace i \in [\NumMeasurements] \mid \ell(i) = k \rbrace} \mathbb{E}_{\Measurement_i} \left[- \frac{\partial^2}{\partial^2 (\PoseSet, \LandmarkSet)} \log p(\Measurement_i \mid \PoseSet, \LandmarkSet) \right]
\end{equation}
for the sum of information matrices from all measurements generated by sensor $k$, equation \eqref{parameterized_SLAM_information_matrix} is equivalent to:
\begin{equation}
\Information(\PoseSet, \LandmarkSet; s) = \sum_{k = 1}^{\NumCandidates} s_k \Information_k(\PoseSet, \LandmarkSet).
\end{equation}
That is: the FIM for the SLAM estimation problem \eqref{parameterized_SLAM_MLE} under sensor arrangement $s$ is simply the sum of the information provided by each individual sensor included in $s$.

\subsection{Performance Criteria for Sensor Arrangements}
%

The CRLB implies that if we want to recover a SLAM estimate $(\Estimate{\PoseSet}(s), \Estimate{\LandmarkSet}(s))$ from \eqref{parameterized_SLAM_MLE} with a  ``small" uncertainty, we must choose a sensor arrangement $s$ such that the corresponding FIM $\Information(\PoseSet, \LandmarkSet; s)$ is as ``large" as possible.  

%
To that end, inspired by \cite{doherty2022spectral}, we propose to use the \emph{E-optimality} criterion as a performance measure for optimizing the design of sensor arrangements.  In brief, this approach requires maximizing $\MinEig(\Information)$, the minimum eigenvalue of $\Information$.  The advantage of E-optimality versus the more common D-optimality (which maximizes $\log \det(\Information)$) is that the latter depends upon the \emph{entire spectrum} of $\Information$, while the former requires only the \emph{single smallest} eigenvalue; this can be calculated very efficiently, even for very large matrices.  Moreover, since $\lambda_1(\Information)$  lower-bounds the spectrum of $\Information$, maximizing this quantity can be interpreted as maximizing a lower bound on $\log \det(\Information)$ itself.
%
%


%
%

We also note that in many SLAM applications, we are mainly interested in the \emph{robot pose estimates} $\Estimate{\PoseSet}$; the landmark locations $\Estimate{\LandmarkSet}$ are only interesting insofar as they support accurate robot localization.  In this case, our main concern is in minimizing $\Cov{\Estimate{\PoseSet}}$, the marginal covariance of the pose estimates.  In light of \eqref{CRLB} and the 2x2 block-matrix inversion formula, the relevant form of the CRLB in this case is:
\begin{equation}
\Cov{\Estimate{\PoseSet}}(s) \succeq \Schur \left( \Information(\PoseSet, \LandmarkSet; s) \right),
\end{equation}
where here $\Schur(\Information)$ denotes the generalized Schur complement of $\Information(\PoseSet, \LandmarkSet; s)$ with respect to the landmark variables $\LandmarkSet$~\cite{li2000extremal}.
%
%
Thus, we propose to use the following objective:
\begin{equation} \label{eq:objective}
\Objective(\PoseSet, \LandmarkSet; s) \Defined \MinEig \left(   \Schur \left( \Information(\PoseSet, \LandmarkSet; s) \right) \right).
\end{equation}

\subsection{Optimal sensor arrangement}
We are now ready to formalize the optimal sensor arrangement problem.  Given a set of candidate sensor mounts $\BaseSet$, a realization of an environment $\LandmarkSet$ and robot trajectory $\PoseSet$, and a number of sensors $\NumCameras$ to select, our task is to find the $\NumCameras$-cardinality subset $S \subset \BaseSet$ that maximizes the objective \eqref{eq:objective}:
\begin{problem}[Optimal Sensor Arrangement] \label{prob:sensor_arrangement}
\begin{equation}
\label{sensor_arrangement}
\begin{aligned}
\Optimal{f}  = \max_{s \in \{0,1\}^N} \ &\SetFunction(\PoseSet, \LandmarkSet; s)  \quad \mathrm{s.t.} \sum_{i=1}^N s_i = K.
\end{aligned}
\end{equation}
\end{problem}

\section{Fast Approximation Algorithms}
In this section we describe fast approximate algorithms for solving the sensor arrangement problem (Problem \ref{prob:sensor_arrangement}).
We resort to approximations because finding the global optimum of \Cref{prob:sensor_arrangement} is NP-hard, and therefore cannot (in general) be solved in polynomial time ~\cite{shamaiah2010greedy}.
Our approach employs both a simple greedy method described in \Cref{sec:greedy} and a convex relaxation developed in \Cref{sec:convex}.  Throughout this section we will drop explicit reference to $\PoseSet$ and $\LandmarkSet$ to ease notation (since these are not variables of optimization).

\subsection{Greedy Maximization} \label{sec:greedy}
As their name suggests, greedy set maximization algorithms iteratively construct a solution set $S \subseteq \BaseSet$ by appending, in each iteration, an element $x \in \BaseSet - S$ that leads to the greatest marginal gain in the value of the objective.  When the objective $\SetFunction$ is \emph{normalized monotone submodular}~\cite{krause_submodular_2014},
the greedy solution $S_{\mathrm{g}} \subset \BaseSet$ is guaranteed to satisfy
\begin{equation}
    \SetFunction(S_{\mathrm{g}}) \geq (1 - \frac{1}{e}) \SetFunction(\Optimal{S}) \approx 0.63 \SetFunction(\Optimal{S}),
\end{equation}
where $\Optimal{S}$ is a globally maximizer of \ref{sensor_arrangement}.
Unfortunately, while $\Objective$ is monotonic and normalized, $\MinEig$ only satisfies \emph{approximate} forms of  submodularity~\cite{chamon2017approximate, hashemi2019submodular}. 

\subsection{Convex Relaxation} \label{sec:convex}
%


Let us relax the (nonconvex) \emph{binary} constraints appearing in \Cref{prob:sensor_arrangement} to (convex) \emph{Boolean} constraints:

\begin{problem}[Boolean Relaxation of \Cref{prob:sensor_arrangement}] \label{prob:convex_relaxation}
\begin{equation}
\label{convex_relaxation}
\begin{aligned}
\Optimal{\mu} = \max_{\omega \in [0,1]^N} \ &\SetFunction(\omega) \quad \mathrm{s.t.} \sum_{i=1}^N \omega_i = K.
\end{aligned}
\end{equation}

%
\end{problem}

Observe that if $\SetFunction$ is concave, then \eqref{convex_relaxation} is a \emph{convex program}.  Consequently, the computational tractability of \Cref{prob:convex_relaxation} hinges on the concavity of our objective function $\SetFunction$.  Fortunately, the following proposition (proven in the Appendix\footnote{ \scriptsize Due to space constraints, the proof and additional figures are provided in the supplement \cite{kaveti2023oasis}}) states that the E-optimality performance criterion $\Objective$ defined in \eqref{eq:objective} is indeed concave. 

\begin{proposition}[Concavity of $\Objective$] \label{prop:concavity}
	The function $\Objective$ defined in \eqref{eq:objective} is concave on the domain $[0, 1]^\NumCandidates$.
\end{proposition}


It follows that Problem \ref{prob:convex_relaxation} is convex when using the E-optimality criterion \eqref{eq:objective}, and thus can be solved \emph{globally optimally} using standard convex optimization methods.  Following the insights of   \cite{doherty2022spectral}, we thus propose to solve \Cref{prob:convex_relaxation} using the Frank-Wolfe method. 


Let us consider how the optimal values of Problems \ref{prob:sensor_arrangement} and \ref{prob:convex_relaxation}  compare.  Since \Cref{prob:convex_relaxation} is a relaxation of \eqref{sensor_arrangement}, its optimal value $\Optimal{\mu}$ provides an upper bound on the optimal value $\Optimal{f}$ of \eqref{sensor_arrangement}.  Conversely, we clearly have $\SetFunction(s) \le \Optimal{\SetFunction}$ for any feasible $s$ in \eqref{sensor_arrangement}.  
These inequalities together imply:
\begin{equation}
\label{suboptimality_bound}
\Optimal{\mu} - f(s) \ge \Optimal{f} - f(s)
\end{equation}
for any feasible $s$ in \eqref{sensor_arrangement}.  The significance of inequality \eqref{suboptimality_bound} is that it enables us to use the optimal value $\Optimal{\mu}$ of Problem \ref{prob:convex_relaxation} (which we can compute efficiently) in order to bound the suboptimality  $\Optimal{f} - f(s)$ of \emph{any} feasible solution $s$ of Problem \ref{prob:sensor_arrangement}.  In particular, as we will see in Section \ref{sec:experiments}, this will provide a practical means of \emph{verifying} the (\emph{global}) optimality of candidate solutions of Problem \ref{prob:sensor_arrangement}.

%

%

\subsection{The OASIS algorithm}

The entire OASIS procedure is summarized in \Cref{alg:oasis}.  In brief, our method applies sequential greedy set maximization to obtain a feasible solution $s_{\mathrm{g}}$ of Problem {\ref{prob:sensor_arrangement}}, and then solves the convex relaxation \Cref{prob:convex_relaxation} (using the Frank-Wolfe method) to obtain an upper bound $\Optimal{\mu}$ on Problem \ref{prob:convex_relaxation}'s optimal value that we can use to bound $s_{\mathrm{g}}$'s suboptimality via \eqref{suboptimality_bound}.  As we show in Section \ref{sec:experiments}, this simple approach enables us to recover certifiably optimal solutions of the sensor arrangement problem {\eqref{sensor_arrangement}} in practice.

%
%
%
\begin{small}
  \begin{algorithm}[t]
    \caption{ OASIS Algorithm} \label{alg:oasis}
    \begin{algorithmic}[1]
      \Input Concave obj.\ function $f$, feasible point $\omega$ of \eqref{convex_relaxation}. 
      \Output A feasible solution $s_{\mathrm{g}}$ of Problem \ref{prob:sensor_arrangement} and an upper bound $\Optimal{\mu} \ge \Optimal{f}$ on Problem \ref{prob:sensor_arrangement}'s optimal value.
      \Function{OASIS}{$\omega$}
        \State $s_{\mathrm{g}} \leftarrow \textsc{Greedy}(\SetFunction)$
        \Comment{Greedy solution; Sec. (\ref{sec:greedy})}
      \State $(\Optimal{\mu}, \Optimal{\omega}) \leftarrow $\textsc{FrankWolfeAC}$(\omega; \SetFunction)$
     \Comment{Solve~\eqref{convex_relaxation}}
      
    
      \State \Return $(s_{\mathrm{g}}, \Optimal{\mu}, \Optimal{\omega})$
      \EndFunction
    \end{algorithmic}
  \end{algorithm}
\end{small}

\captionsetup[subfigure]{skip=3pt, labelfont={scriptsize,bf, sf},font={footnotesize}}

\begin{figure*}[ht!]
\vspace{1mm}
\centering
\captionsetup{font={footnotesize}, labelfont={bf,sf}}
\begin{tabular}{c}

\subfloat[]{\includegraphics[clip, trim={0cm 0.6cm 0.5cm 0.5cm},width=0.242\linewidth]{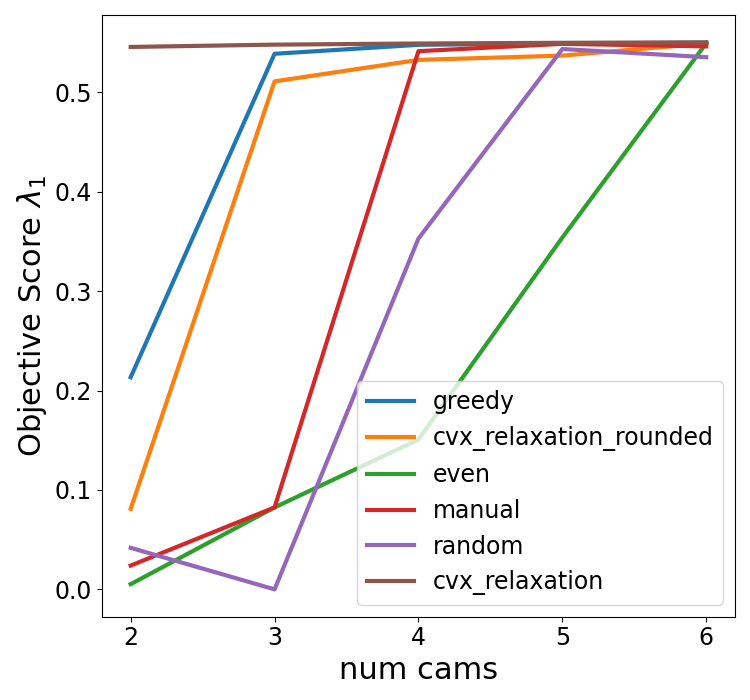}}
\subfloat[]{\includegraphics[clip, trim={0cm 0.6cm 0.5cm 0.5cm},width=0.242\linewidth]{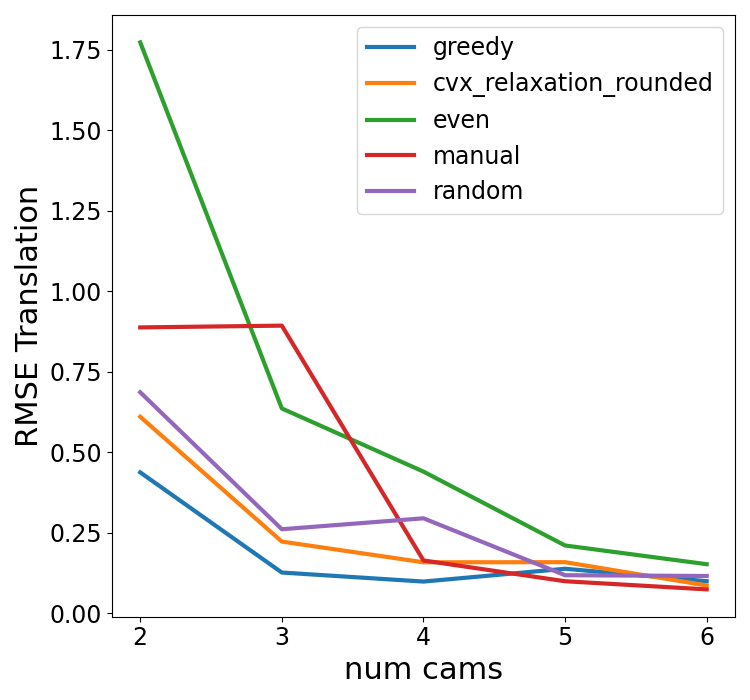}}
\subfloat[]{\includegraphics[clip, trim={0cm 0.6cm 0.5cm 0.5cm},width=0.242\linewidth]{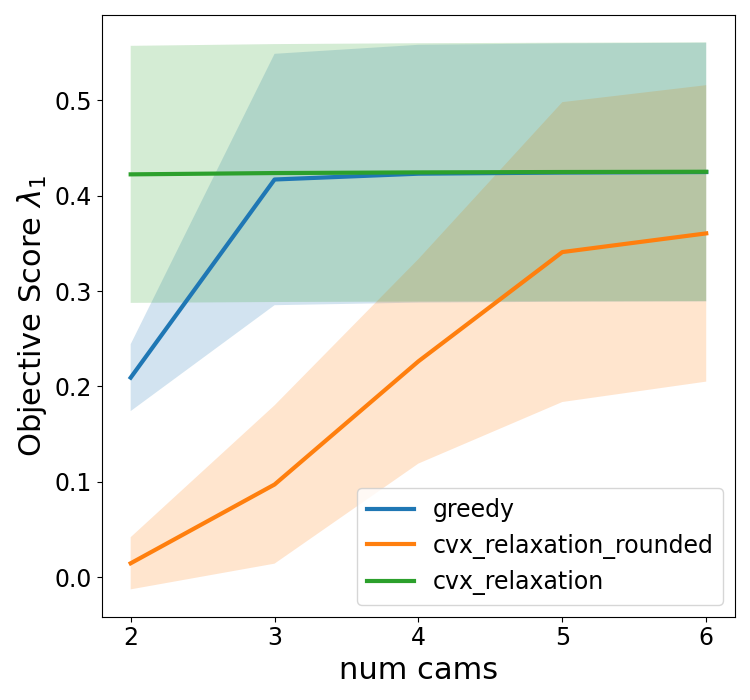}}
\subfloat[]{\includegraphics[clip, trim={0cm 0.6cm 0.5cm 0.5cm},width=0.245\linewidth]{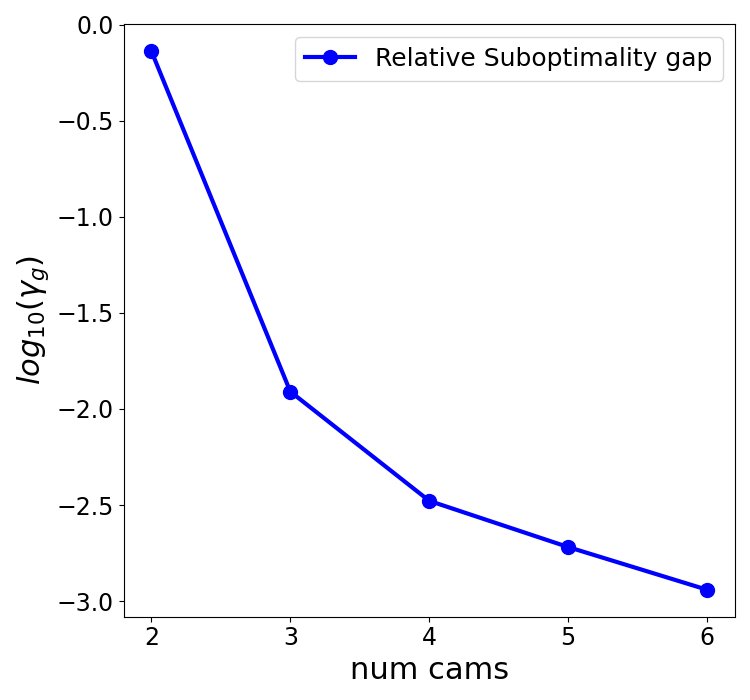}}

\end{tabular}
\caption{\textbf{Experimental Results I:} Quantitative results of optimal placement on the synthetic dataset of 50 simulations of a random motion of the sensor rig with varying number of selected cameras, $k$. (a) and (b) Show the trend of the optimized score of the objective function $\lambda_1$ and median RMSE of the translational component of the pose estimates computed from the graph resulting from the optimal camera selection with respect to the ground truth computed across simulations. (c) Gives a closer look at the mean and standard deviation of the objective scores of greedy, convex relaxation solutions before and after k-max rounding across simulations. The greedy optimization results in a near-optimal solution, as demonstrated by its closeness to the score of the unrounded convex relaxation approach, which gives the upper bound on the optimal value, especially for $k > 2$. (d) Shows that we achieve a very tight relative sub-optimality gap, $\gamma^* = f(w^*) - x_g/x_g$ asserting the effectiveness of submodular greedy optimization.}
\label{fig:numcams_score_rmse}
\vspace{-2mm}
\end{figure*}
\begin{figure*}[ht!]
\centering
\captionsetup{font={footnotesize}, labelfont={bf,sf}}
\begin{tabular}{c}
\subfloat[]{\includegraphics[clip, trim={0cm 0.4cm 0.3cm 1.5cm},width=0.23\linewidth]{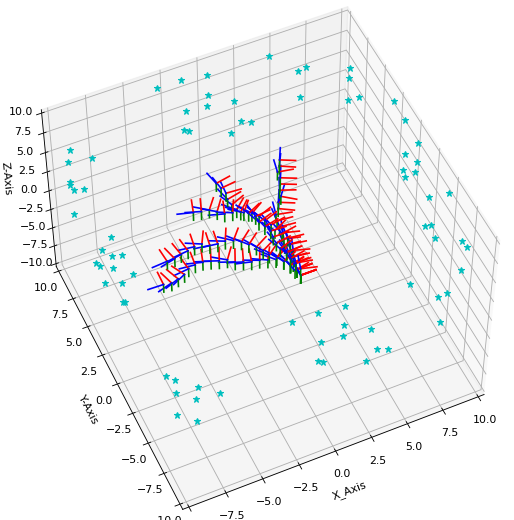}}
\subfloat[]{\includegraphics[clip, trim={0cm 0.4cm 0.3cm 1.5cm},width=0.23\linewidth]{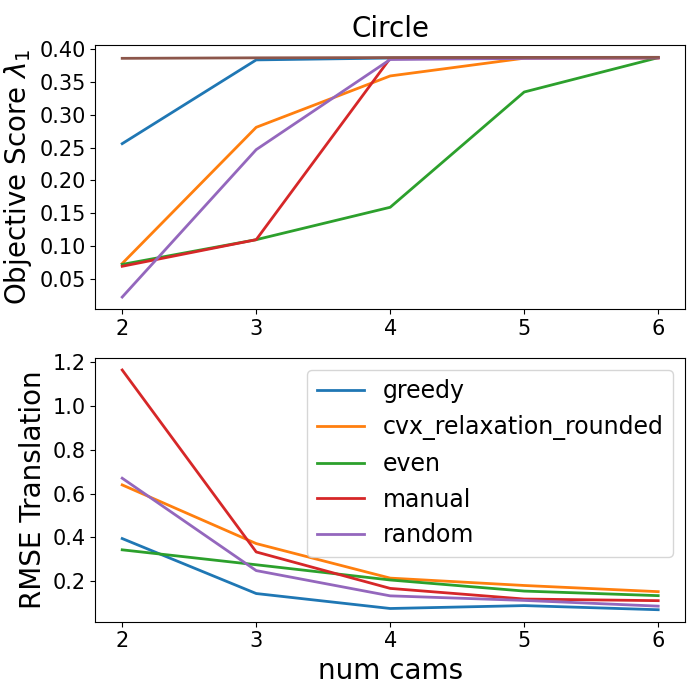}}
\subfloat[]{\includegraphics[clip, trim={0cm 0.4cm 0.3cm 1.5cm},width=0.23\linewidth]{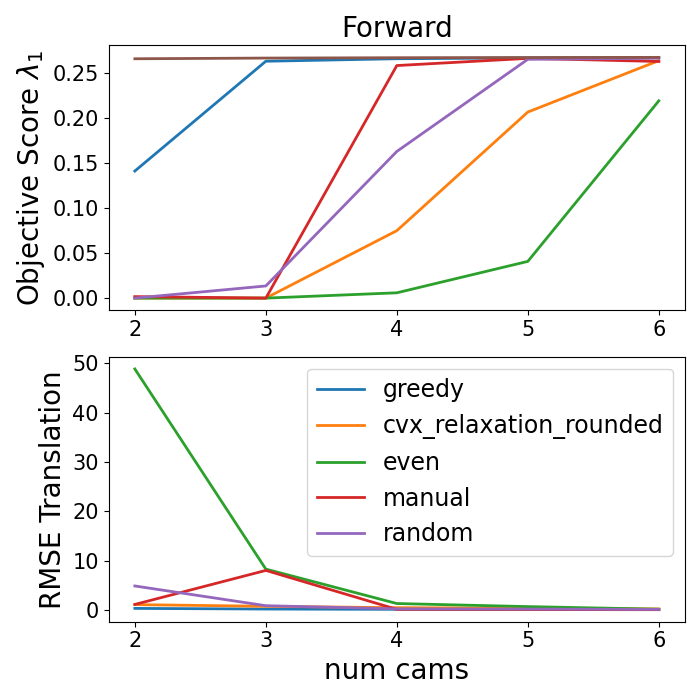}}
\subfloat[]{\includegraphics[clip, trim={0cm 0.4cm 0.3cm 1.5cm},width=0.23\linewidth]{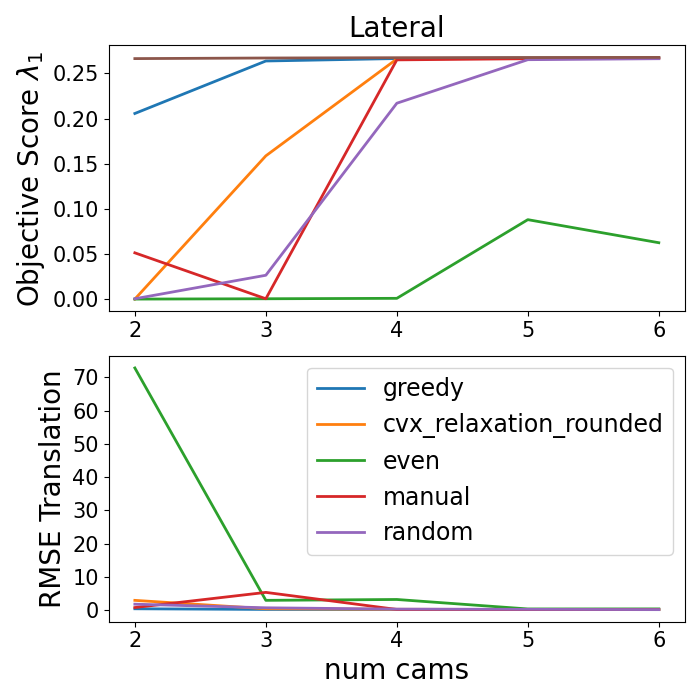}}
\end{tabular}
\caption{\textbf{Experimental Results II:} (a) The synthetic data collection setup. A simulated room-like environment with landmarks and random trajectories from the top view. (b-d) 
 Quantitative results showing median score and RMSE of the benchmarking algorithms for Circular, Forward and Lateral motions.}
\label{fig:dummy}
\vspace{-2mm}
\end{figure*}

\newsavebox{\measurebox}
\begin{figure*}
\centering
\captionsetup{font={footnotesize}, labelfont={bf,sf}}
  \begin{minipage}[b]{.28\textwidth}
    \subfloat[]{\label{fig:figA}\includegraphics[width=\textwidth, height=4cm]{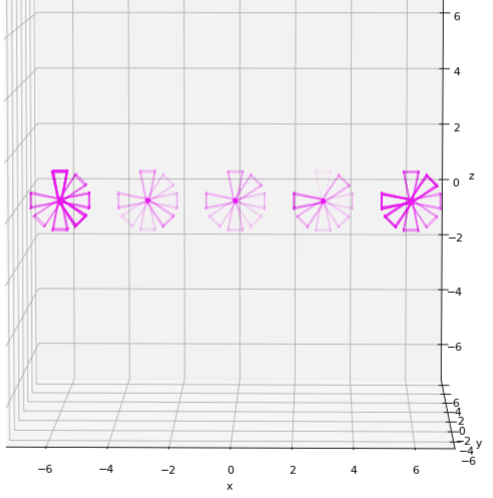}}
  \end{minipage}
  \sbox{\measurebox}{%
  \begin{minipage}[b]{.28\textwidth}
  \subfloat[]{\label{fig:figD}\includegraphics[width=\textwidth,clip, trim={0cm 0cm 0cm 0.5cm},height=4cm]{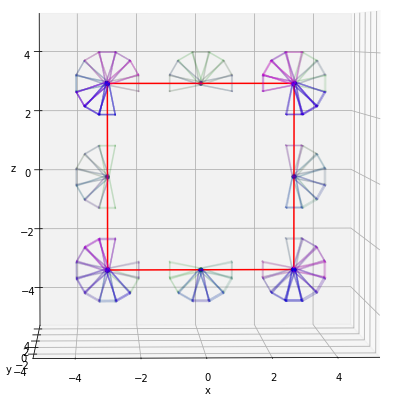}}
  \end{minipage}}
\usebox{\measurebox}\qquad
\begin{minipage}[b][\ht\measurebox][s]{.26\textwidth}
\centering
\subfloat[]{\label{fig:figB}\includegraphics[width=\textwidth,clip, trim={0cm 0cm 0cm 0cm}, height=1.9cm]{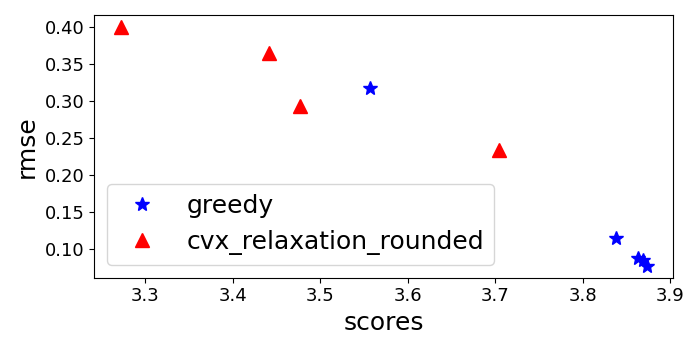}}\\
\subfloat[]{\label{fig:figC}\includegraphics[width=\textwidth,clip, trim={0cm 0cm 0cm 0cm}, height=1.9cm]{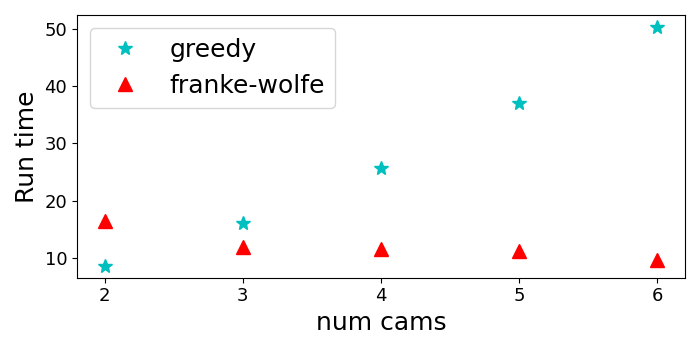}}
\end{minipage}
\caption{\textbf{Experimental Results III.} Visualization of the greedy optimal selection across multiple experiments overlayed for a candidate pool (a) lying on a linear array configuration and (b) regularly spaced in both translation and orientation. Deeper/darker colors indicate a higher frequency of selection. (c) Shows that the score $\lambda_1$ is inversely related to the RMSE of SLAM pose estimates for both greedy and convex relaxation approaches. Thus, E-optimality improves SLAM performance. (d) Run time comparison of greedy optimization with the convex relaxation approach. Time complexity of the greedy method increases linearly with the number of cameras, while there is not much effect on the convex relaxation approach.}
\label{fig:hot_spots}
\vspace{-5mm}
\end{figure*}

\section{Evaluation} \label{sec:experiments}
Here we describe our experimental evaluation, and present results of greedy and convex optimization approaches in terms of both the information-theoretic criterion $\Objective$ in \eqref{eq:objective} and downstream SLAM performance. 
\subsection{Experimental setup} \label{sec:experiments_a}
\textbf{Datasets} We employ landmark-based visual SLAM as the evaluation framework for our proposed optimization methods. We developed a Python-based simulation system \footref{note1} tailored to generate different synthetic datasets for assessing the performance of various sensor arrangement methods(shown in \cref{fig:dummy}(a)). Within this simulated environment, a room-like setting was constructed, housing a mobile multi-camera sensor rig programmed to exhibit diverse motion patterns, including circular movements, straight-line forward and lateral motions, and random movements. The simulated environment offers several advantages, including the capacity to conduct a large number of reproducible experiments with fixed parameters related to various aspects such as the number of poses, camera fields of view (FoVs), landmark distribution, environmental characteristics, rig motion patterns, among others. Moreover, it provides support for incremental development and debugging.
In our simulations, we generate four distinct trajectory types: circular, lateral, forward, and random motions, using a simulated sensor rig equipped with 68 different candidate camera poses regularly spaced on a square frame as shown in \cref{fig:hot_spots}(b). We aggregate the results from 50 simulations for each trajectory type, with each simulation involving varying landmark distribution. 

\textbf{Benchmark Algorithms}
In all experiments, we compare our algorithms with the following benchmarks:
\begin{enumerate}
    \item random sampling from the candidate pool (random);
    \item a standard configuration from a similar commercially available autonomous vehicle (manual);
    \item a ``uniform coverage" heuristic which evenly spaces the cameras in terms of viewing direction and position throughout the vehicle's feasible set of $\LieGroupSE{3}$ (even);
\end{enumerate}
These heuristic-based methods act as baselines to demonstrate the effectiveness of set function maximization for the camera placement problem.

\textbf{Metrics} For each dataset, we consider a selection of k=2,3...6 cameras from the candidate pool for the optimal camera placement problem. We evaluate each method by comparing the score of the objective function $\Objective$ in \eqref{eq:objective}, and the results of the maximum likelihood estimation (MLE) of the SLAM problem for the camera sets selected by respective methods. To evaluate SLAM performance, we compute the  Root Mean Square Error (RMSE) between the MLE estimates of poses obtained by performing optimization (using GTSAM \cite{Dellaert2012GTSAM}) on the factor graph induced by camera selection and the ground truth poses. We also assess the performance of greedy and convex relaxation approaches in terms of run-time and ability to produce optimal solutions. 

\subsection{Results}
\subsubsection{Comparison with benchmarks} Fig.~\ref{fig:numcams_score_rmse} and ~\ref{fig:dummy} provides a concise summary of quantitative outcomes from the benchmarking algorithms discussed in Section \ref{sec:experiments_a}, applied to various camera selections. It is evident that our proposed greedy optimization consistently outperforms the even, manual, and random selection methods. This performance gap becomes more pronounced when dealing with a smaller number of cameras to be positioned (k). This observation underscores the effectiveness of the set function maximization approach for achieving optimal sensor placement.

Fig.~\ref{fig:hot_spots}(c) illustrates the relationship between the value of the objective $\Objective(s)$ for our optimized sensor arrangements $s$, and the empirical RMSE of the SLAM estimates $\Estimate{X}(s)$ obtained using that design within our experiments. These results demonstrate that improving the value of our proposed information-theoretic criterion $\Objective$ in \eqref{eq:objective} is indeed predictive of improved downstream SLAM performance.

\subsubsection{Greedy vs Convex Relaxation} Fig.~\ref{fig:numcams_score_rmse} compares the objective value achieved by the greedy method with (i) the upper bound $\Optimal{\mu} \ge \Optimal{f}$ on Problem \ref{prob:sensor_arrangement}'s optimal value provided by the Boolean relaxation \eqref{convex_relaxation}, and (ii) the objective value obtained by \emph{rounding} the solution $\Optimal{\omega}$ of the convex relaxation using $K$-max rounding.

These results reveal several very interesting trends.  First, we observe that the upper bound $\Optimal{\mu}$ on Problem \ref{prob:sensor_arrangement}'s optimal value $\Optimal{f}$ provided by the convex relaxation and the lower bound $\Optimal{f} \ge f(s_{\textnormal{g}})$ provided by the greedy solution are remarkably close across the majority of our experiments.  In light of inequality \eqref{suboptimality_bound}, this shows that (i) the heuristic greedy method is remarkably effective in finding high-quality sensor arrangements in these experiments, and (ii) the convex relaxation's optimal value $\Optimal{\mu}$ is a remarkably sharp approximation of Problem \ref{prob:sensor_arrangement}'s optimal value, enabling us to \emph{certify} the optimality of the greedy method's solutions.  Indeed, examining Fig.~\ref{fig:numcams_score_rmse}(d), which plots the upper bound $(\Optimal{\mu} - f(s_{\textnormal{g}})) / f(s_{\textnormal{g}}) \ge (\Optimal{\mu} - f(s_{\textnormal{g}})) / \Optimal{f}$ on $s_{\textnormal{g}}$'s \emph{relative} suboptimality, reveals that the greedy method succeeds in finding solutions that are  (at most)  $1-2\%$ suboptimal in the majority of cases.  (Note that this suboptimality bound is a substantial improvement on the usual $1-1/e$ suboptimality bound obtained from greedy submodular maximization). At the same time, however, the sensor arrangements recovered by rounding the actual \emph{solutions} $\Optimal{\omega}$ returned by Problem \ref{prob:convex_relaxation} are consistently significantly suboptimal; evidently while the \emph{objective value} of Problem \ref{prob:convex_relaxation}'s maximizers very tightly approximates $\Optimal{f}$, the solutions $\Optimal{\omega}$ are not close to  integral.   

Altogether these results show that while greedy maximization or convex relaxation would not \emph{individually} suffice to obtain certifiably optimal solutions, their combination is remarkably effective in practice. Finally, we note that one limitation of the greedy method is that its per-iteration complexity is linear in $\NumCandidates$ (see \cref{fig:hot_spots}(d)). This However, this is not a significant hindrance for sensor rig design, as it can be conducted as an offline task. 

\subsubsection{Qualitative Analysis} Figures \ref{fig:hot_spots}(a) and (b) showcase the qualitative outcomes of the greedy optimization approach. In these visualizations, we have superimposed the candidate pool and the camera selections optimized through the greedy approach across all experimental runs. This graphical representation aids us in identifying hot spots or key areas of interest for camera placement. In Figure \ref{fig:hot_spots}(a), we depict the results of selecting two cameras from a candidate set arranged linearly. Notably, the extremities of the linear array are consistently favored. This observation aligns with the preference for a larger baseline, a characteristic advantageous when utilizing two cameras. Turning our attention to Fig.~\ref{fig:hot_spots}(b), we consider the scenario of positioning cameras on a mobile robot. Here, we observe a notable preference for the four extreme corners over other potential positions.

\section{Conclusion} \label{sec:conclusion}

In this paper we proposed a formal methodology for optimally arranging the sensors on a mobile robot designed to perform SLAM.  Our approach formalizes the design task as an optimal subset selection problem under a computationally tractable E-optimality performance measure. While subset selection problems are NP-hard in general, we also developed a fast approximate optimization scheme that combines greedy sensor selection with convex-relaxation-based post-hoc suboptimality bounds. Our experimental evaluations show that our approach is remarkably effective in practice, enabling the efficient recovery of sensor arrangements that are within 1-2\% of the optimal value.

While we have currently focused on camera selection for visual SLAM, our future work will explore designing a broader class of sensor rigs with heterogeneous sensors, incorporating a richer set of design constraints (SWAPC) than cardinality, and conducting real-world experiments.

\section*{APPENDIX}
\subsection{Proof of \Cref{prop:concavity}}
First, note that $\Information(\omega) \succeq 0$ for all $\omega \in [0, 1]^K$ by construction.
Our proof relies on four key properties of $\Objective = \lambda_1\circ\Schur\circ\Information$:
\begin{enumerate}
	\item $\Information(\cdot)$ is an affine function of its argument;
	\item the Schur complement $\Schur(\cdot)$ is matrix concave on $\PSDMatrices{n}$~\cite{li2000extremal};
	\item the minimum eigenvalue function $\lambda_1(\cdot)$ is monotonic with respect to the Loewner order ($\succeq$)~\cite{horn2012matrix}; and
	\item $\lambda_1(\cdot)$ is concave over $\SymmetricMatrices{n}$~\cite{rockafellar1997convex}.
\end{enumerate}
We have, for all $\theta \in [0,1]$ and any $\omega, \nu \in [0,1]^K$: 
\begin{equation}
\begin{aligned}
	\Objective(\theta \omega + (1-\theta)\nu) &= (\lambda_1 \circ \Schur)(\Information(\theta\omega + (1-\theta)\nu)) \\ 
	{\color{gray} \text{(Property 1) }} &= (\lambda_1 \circ \Schur)(\theta\Information(\omega) + (1-\theta)\Information(\nu)) \\ 
	{\color{gray} \text{(Properties 2 \& 3) }} &\geq \lambda_1(\theta \Schur(\Information(\omega)) \ + \\ 
	&\enspace (1-\theta)\Schur(\Information(\nu))) \\
	{\color{gray} \text{(Property 4) }}&\geq \theta\lambda_1(\Schur(\Information(\omega))) \ + \\ 
	&\enspace (1-\theta)\lambda_1(\Schur(\Information(\nu))) \\
	& = \theta \Objective(\omega) + (1-\theta)\Objective(\nu),
\end{aligned}
\end{equation}
which proves concavity as desired.
	
\bibliographystyle{IEEEtran}
\bibliography{references}
\end{document}